\documentclass[10pt, a4paper]{article}

\usepackage{lrec2026} 

\usepackage{amsmath}
\usepackage{booktabs}
\usepackage{ragged2e}
\usepackage{multirow}
\usepackage{tcolorbox}
\usepackage{subcaption}

\title{Conversational Control with Ontologies for Large Language Models:\\
A Lightweight Framework for Constrained Generation}

\name{Barbara Gendron$^1$, Gaël Guibon$^{1,2}$, Mathieu d'Aquin$^1$} 

\address{$^1$Université de Lorraine, CNRS, LORIA, Nancy, France \\
         $^2$Université Sorbonne Paris Nord, LIPN, CNRS, UMR 7030, F-93430, Villetaneuse, France \\
         \{barbara.gendron, gael.guibon, mathieu.daquin\}@loria.fr\\}

\abstract{
Conversational agents based on Large Language Models (LLMs) have recently emerged as powerful tools for human-computer interaction. Nevertheless, their black-box nature implies challenges in predictability and a lack of personalization, both of which can be addressed by controlled generation. This work proposes an end-to-end method to obtain modular and explainable control over LLM outputs through ontological definitions of aspects related to the conversation. Key aspects are modeled and used as constraints; we then further fine-tune the LLM to generate content accordingly. To validate our approach, we explore two tasks that tackle two key conversational aspects: the English proficiency level and the polarity profile of the content. Using a hybrid fine-tuning procedure on seven state-of-the-art, open-weight conversational LLMs, we show that our method consistently outperforms pre-trained baselines, even on smaller models. Beyond quantitative gains, the framework remains model-agnostic, lightweight, and interpretable, enabling reusable control strategies that can be extended to new domains and interaction goals. This approach enhances alignment with strategy instructions and demonstrates the effectiveness of ontology-driven control in conversational systems.
 \\ \newline \Keywords{conversation ontology, large language models, fine-tuning} }

\begin{document}

\maketitleabstract

\section{Introduction}
\label{sec:intro}

Conversational agents based on Large Language Models (LLMs) have become increasingly present in everyday life, raising questions about the need for more controlled and predictable interactions~\cite{hennekeuser2024don}. Although LLMs exhibit impressive generative abilities due to training on massive corpora~\cite{chiang-etal-2022-recent}, their black-box nature hinders the assessment of their suitability for goal-oriented and domain-specific dialogue~\cite{bellos-etal-2024-large}. This motivates the growing interest in knowledge-enhanced conversational agents~\cite{erickson2025llm}, especially in applications such as customer support~\cite{su-etal-2025-llm}, healthcare~\cite{cho-etal-2023-integrative,liu-etal-2025-interactive}, and human resources~\cite{xu-etal-2024-hr}. In such use-cases, interactions are context-dependent, and the user is seeking meaningful answers, which implies predictable outputs from the agent. This need for external control is not limited to domain-specific agents. Even in open-domain dialog, the ability to guide content generation is now acknowledged as critical, as evidenced by the recent initiative of the first workshop on simulating conversational intelligence~\cite{graham-etal-2024-findings}. Despite recent advances in constrained generation~\cite{su-etal-2021-plan-generate}, many of these approaches are costly and model-dependent. Moreover, LLMs struggle to satisfy complex or multi-dimensional constraints in a consistent manner~\cite{sun-etal-2023-evaluating}.

To solve these limitations, we propose a model-agnostic framework for conversational control through a lightweight fine-tuning procedure that enables predictable, descriptor-driven generation. Our approach involves the definition of descriptors that characterize utterances and a strategy that governs the evolution of descriptor values during the conversation. While descriptors offer a static representation of utterance properties, the strategy enables the dynamic adaptation of the agent. To support reliable and interpretable control, we formalize descriptor definitions within an ontology. It enables a structured, logic-based representation of concepts and their relationships~\cite{Gruber1993ATA, ontoref}. Ontologies allow for consistent definitions, explicit reasoning, and alignment with human knowledge, which is essential for designing agents that exhibit transparent and reproducible behavior. Indeed,~\citet{varshney:2023} emphasize that agents aware of human features improve user experience by facilitating meaningful dialogs that recognize and respond to emotions. The knowledge-driven foundation of our approach facilitates knowledge engineering at both the utterance level (via descriptor annotation) and the conversation level (via the descriptor evolution strategy throughout the conversation). Ultimately, it enables the integration of external knowledge into LLM-based systems~\cite{Pan_2024} without altering the underlying model architecture.

Therefore, we present an ontology-based framework for controlled conversation generation in LLMs, incorporating structured knowledge for adaptive and predictable outputs. We address the research question: \emph{How can knowledge from ontological definitions be processed to control the generation of a conversational LLM?} We further explore how relevant descriptors can be selected and modeled from user content across different aspects of the conversation.

To incorporate such knowledge in an LLM using constrained generation, we develop a fine-tuning procedure with the objective of improving generation compliance with the conversation strategy instructions. We illustrate our approach using two use-cases: \textbf{Proficiency-Level Control}, which consists of adapting the English language level of the agent to what has been previously detected as understandable by the user, and \textbf{Polarity Profile Control}, which consists of presenting positive, negative, and neutral content that can be emotionally loaded or not, depending on the emotion detected in the user's prompt. For each use-case, we identify descriptors to define in an ontology. Utterances are assigned to ontological classes based on descriptor values. The conversation strategy is then defined in the ontology from these classes, enabling ontological reasoning to determine the class of the next utterance. To effectively apply conversational control, we fine-tune several LLMs on constrained generation with respect to all possible classes. Finally, we evaluate the generated content in terms of compliance with both the utterance classes and the conversation strategy. Our contribution is two-fold:

\begin{itemize}
    \item We propose a novel methodology to fine-tune LLMs using ontological definitions that enable controlled generation, allowing the guidance of a conversational agent through ontologically-defined strategies. 
    \item Using 2 use-cases, our approach yields enhanced controlled generation, both quantitatively and qualitatively.
\end{itemize}

For the sake of reproducibility, we publicly share our code, data, and ontologies\footnote{\url{https://github.com/B-Gendron/OntoCLMConvControl}}.

\section{Related Work}
\label{sec:sota}

\subsection{Knowledge-Driven Language Modeling}

Most of the contributions about unifying LLMs and knowledge-based systems, such as ontologies and knowledge graphs (KGs), focus on improving knowledge engineering thanks to language modeling. Regarding ontologies specifically, work has recently been directed toward ontology alignment~\cite{he2023exploring} and ontology learning~\cite{giglou:2023}. Our work pursues the opposite objective, which is to use ontology to improve LLM output. This can be performed by following several objectives.~\citet{agrawal-etal-2024-knowledge} study the efficiency of such methods to mitigate hallucinations inherent to LLMs~\cite{hallu}. KGs can also be used to extend the internal representation of LLM to structured knowledge~\cite{perozzi2024letgraphtalkingencoding}. Finally, the field of application for this hybridization has been the most studied in its ability to assist in specific, interactive, and increasingly complex tasks. These range from Graph QA~\cite{fatemi:2023}, Knowledge-Grounded QA~\cite{sun-etal-2022-jointlk, Wu2023RetrieveRewriteAnswerAK} to Domain Adaptation tasks~\cite{sreedhar-parisien-2022-prompt}. Furthermore, \citet{gloria-silva-etal-2024-plan} demonstrated that hybridization contributes significantly to planning tasks. As conversational tasks are interactive and complex, such a hybridization setup is suitable for dialog systems~\cite{kang2023knowledgegraphaugmentedlanguagemodels}. Particularly,~\citet{varshney:2023} highlight the relevance of understanding human emotions for conversational models, claiming that an agent with emotional awareness enhances user experience by engaging in a meaningful dialog that acknowledges emotions. In this paper, we propose a hybrid LLM/ontology framework where LLM outputs are guided by an ontology-defined conversation strategy, ensuring a consistent and predictable conversational flow focused on proficiency or polarity aspects. Moreover, our pipeline is designed to facilitate the expression of conversation control strategies in a way that is simpler than the intricate hybridization architectures leveraging numerous different modules~\cite{Varshney2023-emokbgan}.

\begin{figure*}[t]
    \centering
    \includegraphics[width=\linewidth]{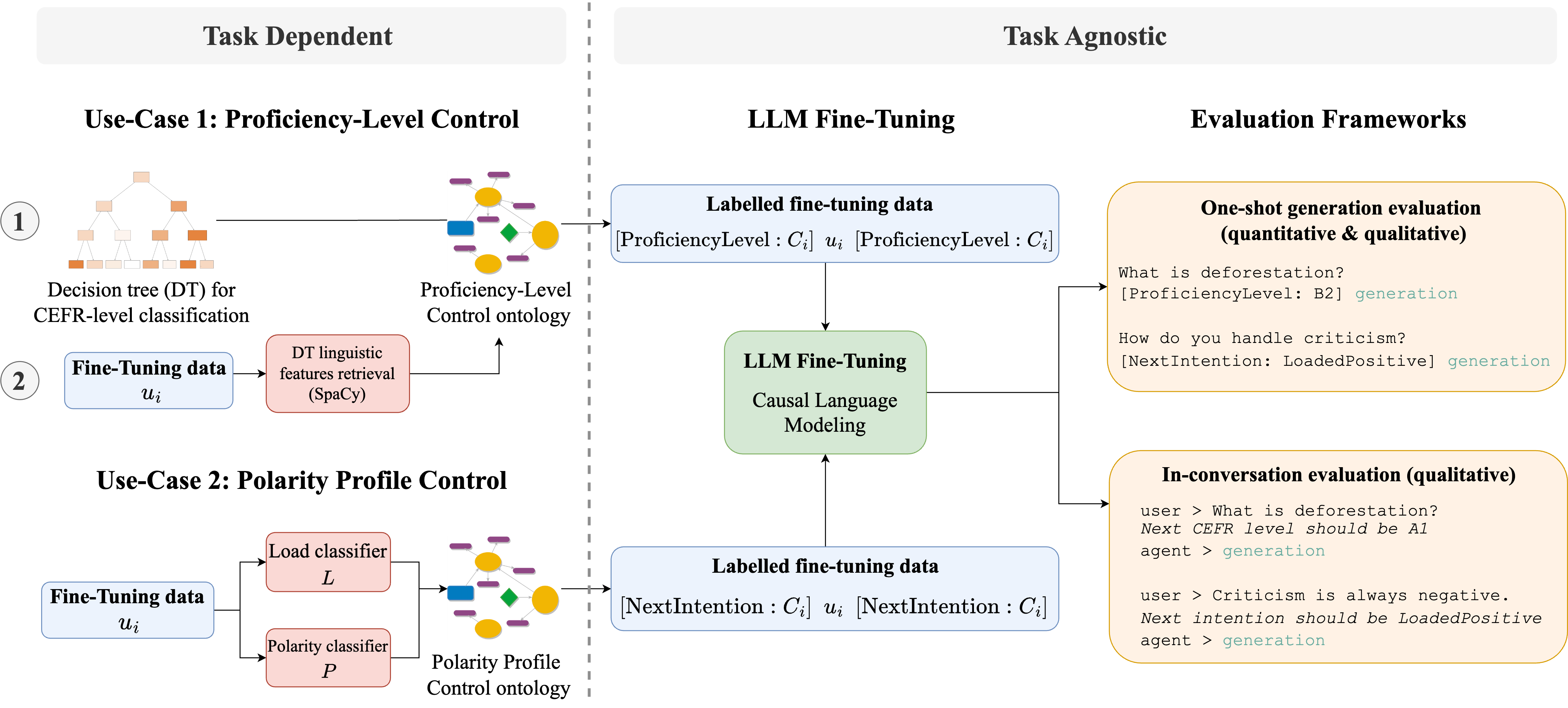}
    \caption{\centering The proposed approach applied to both use-cases. Proficiency-Level Control involves a two-step process: first, quantitative CEFR criteria are obtained from the decision tree output; then, the ontology can be built.}
    \label{fig:all}
\end{figure*}

\subsection{Conversational Control}

Constrained textual generation consists of estimating $p_\theta(x|c)$ instead of simply the probability $p_\theta(x)$ of the token $x$ in a model with parameters $\theta$, where $c$ is the constraint expression. There are various ways to implement a constraint; the most intuitively straightforward method is to concatenate a control code to the prompt to guide generation~\cite{hu2017toward,surveycontrol}. This control code is typically distinguished from the actual prompt content using separators such as brackets. 
Additionally, its presence can be explicitly indicated in a system prompt during fine-tuning.

This approach has been proven efficient for constrained generation, whether this control code is an inherent part of the input~\cite{keskar:2019,goswamy2020adapting} or is handled in a separate dedicated module~\cite{chan2021cocon,Zhang_Wang_Li_Ao_He_2025}. It has been shown that the addition of a control code is a costly option in full fine-tuning scenarios~\cite{chaffin-etal-2022-ppl}, but this limitation can be overcome by using LoRA adapters~\cite{hu2022lora} which is a lightweight yet efficient approach to fine-tuning. This has been done for learning applications to control the generation over selected grammar rules~\cite{glandorf-etal-2025-grammar}. In another direction, recent work further highlights dialog-oriented control and evaluation using LLMs to guide the conversation -- thus leveraging LLM-as-a-judge paradigm~\cite{li-etal-2025-generation} -- for instance applied to autobiographic interviewing~\cite{duan-etal-2025-guidellm}. Nevertheless, in our case, as we want to leverage an hybrid setup, we prefer the control code framework where such code is determined using inference of the definitions of the ontology classes, that are written in description logic. This extension enables controlling generation across hierarchical and abstract levels using the same training procedure. This provides a more expressive naming that fully leverages the expressivity of ontologies.

Finally, we focus on the specific aspects that these methods aim to control in the generation process, as reviewed in~\cite{liang2024controllabletextgenerationlarge}. Controlled generation applies to both content and attributes in generated text. Content control ensures compliance with structural and vocabulary rules through predefined formats such as recipes~\cite{liu-etal-2022-plug}, thus maintaining clarity with organized paragraphs, headings, and adapted lengths~\cite{hua-wang-2020-pair}. Attribute control focuses on higher-level traits such as sentiment, style, and topic. This includes creating text with specific emotional tones or adapting writing styles to domain-specific needs. For example,~\citet{krause-etal-2021-gedi-generative} present a control-code based method to reduce toxicity while maintaining relevance. Related to one of our use-cases,~\citet{malik-etal-2024-tarzan} present a proficiency-level control task. They propose a regression approach for Proficiency level prediction integrated into a Proximal Policy Optimization --~PPO~\cite{PPO}~-- pipeline to fine-tune an LLM. This work leverages a GPT-4 distillation and therefore is a closed-source approach. Ours is open-source and relies on open-weight LLMs.

\begin{figure*}[!ht]
    \centering
    \begin{subfigure}[b]{0.5\linewidth}
        \centering
        \includegraphics[width=0.85\linewidth]{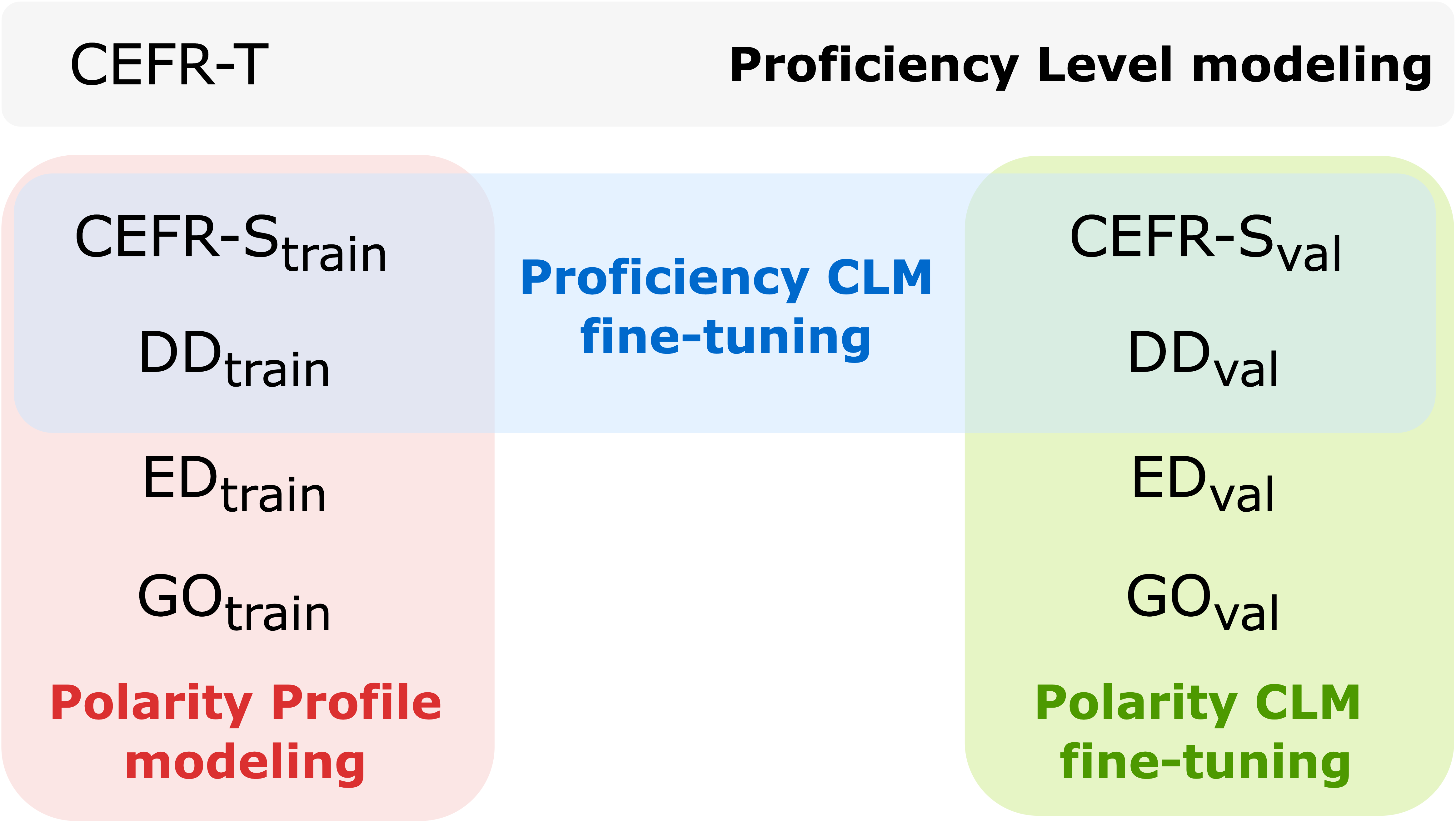}
        \caption{\centering A visualization of the data sources. DD: DailyDialog, ED: EmpatheticDialogues, GO: GoEmotions.}
    \end{subfigure}\hfill
    \begin{subtable}[b]{0.5\linewidth}
        \centering
        \begin{tabular}{@{}lll@{}}
        \toprule
        \textbf{Name (Original Dataset)} & \textbf{\#Utt.} & \textbf{Type} \\ \midrule
        CEFR-T (CNN/DailyMail) & 1,499        & News            \\
        CEFR-S  (CEFR-SP)    &  10,004       & Phrases       \\
        DDBal (DailyDialog) &  37,415       & Dialogs       \\
        DailyDialog         &  102,903      & Dialogs       \\
        GoEmotions          &  54,260       & Posts \\
        EmpatheticDialogues & 18,889       & Dialogs       \\
        SST-3  (SST-5)     &   11,855       & Reviews   \\ \bottomrule
        \end{tabular}
        \caption{\centering Dataset description for both tasks. \#Utt. counts the utterances. All the content is in English.}
    \end{subtable}
    \caption{\centering A description of the data sources used in both use-cases.}
    \label{fig:data}
\end{figure*}

\section{Methodology}
\label{sec:methodo}

This section describes the methodology employed for both use-cases, giving some insight into how we integrate textual descriptors into ontologies so that the corresponding ontology classes can be used in the conversation control strategy. We also define both the training and evaluation setups. An overview of the approach is given in Figure~\ref{fig:all}.

\paragraph{Definitions and Strategy.} We identify the conversational aspects to model based on the intended type of control. For the remainder of this section, we assume that the relevant descriptors have already been identified and defined within an ontology. The next section explains how these descriptors were selected and how their corresponding ontology classes were defined for each use-case. From the ontological description of the aspects we want to control in the conversation, we implement a strategy to guide the conversational flow in a way that remains natural and useful for the user. This is achieved by inferring the ontology class of the next utterance to be generated. In our approach, we build the ontology in Protégé~\cite{protege} and inference is made using the latest \texttt{Pellet}\footnote{\url{https://github.com/stardog-union/pellet}} present in the \texttt{owlready2}\footnote{\url{https://owlready2.readthedocs.io/en/v0.47/}} Python library. Therefore, we control the generation of the next utterance according to the inferred ontology instruction.

\paragraph{Generation Fine-Tuning.} Conversational control is obtained both from the definition of the conversation strategy, as explained above, and from generation control. We find that the latter can be achieved thanks to Causal Language Modeling (CLM) fine-tuning on labeled data. This implies collecting utterances from appropriate data sources (see the next section for dataset descriptions), evaluating the descriptor values of the utterances, and inferring the corresponding ontology class from them. Afterwards, to highlight the utterance's affiliation to a certain class, we enclose this information between brackets on both sides of the utterance, in the form $[o.C: C_i] \enspace u_j \enspace [o.C: C_i]$, where $o$ is the ontology, $o.C$ is a concept from $o$, $C_i$ is the $i^{\text{th}}$ subclass of $C$ in $o$, and $u_j$ is the utterance from the $j^{\text{th}}$ data sample. In the following, we denote the above pattern as a \textit{label-wrapped data sample} (or a label-wrapped utterance). We perform pre- and post-utterance labeling on fine-tuning data to link ontology concept tokens to their most probable counterparts: pre-utterance labels guide generation, while post-utterance labels ensure alignment and control towards the current label. For the Proficiency-Level Control use-case, we evaluated various labeling strategies and selected label wrapping as the optimal approach. Eventually, we fine-tune several LLMs based on Llama3~\cite{dubey2024llama3herdmodels}, Qwen~\cite{yang2024qwen2technicalreport}, Phi~\cite{abdin2024phi3technicalreporthighly}, Mistral~\cite{jiang2023mistral7b}, and a distilled version of DeepSeek-R1~\cite{2025Natur.645..633G} fetched from the \texttt{transformers} HuggingFace library\footnote{\url{https://huggingface.co/docs/transformers/}}. The fine-tuning process uses LoRA (Low-Rank Adaptation) adapters~\cite{hu2022lora, pmlr-v97-houlsby19a} in each decoder layer. This approach significantly reduces the number of trainable parameters, allowing for efficient adaptation to new tasks without requiring extensive computational resources.

\paragraph{Evaluation.} As the conversation strategy is defined consistently in the ontology, it is guaranteed that the right label will be asked at the right time. Therefore, what we need to assess at evaluation time is the model ability to actually generate content according to the requested label. We call this evaluation process \textit{zero-shot generation}, which means that we provide a question to engage the generation, followed by the left-hand bracketed part of our labeling form. This content is included in a system prompt to prevent unwanted pathological behaviors. We then post-process the output to remove the labeling form's right part, which is often generated after fine-tuning. Finally, ontology inference is performed on the generated content to determine its actual label, making our evaluation process similar to those used for classification tasks. Thus, typical classification metrics are used (F1 score and Accuracy), along with one specific metric per use case:  for Proficiency-Level Control, scaled labels necessitate the use of Mean Absolute Error --~MAE~\cite{mae:2005}~-- due to the ordinal nature of CEFR levels. For Polarity Profile Control, prediction relevance is evaluated using the Matthews Correlation Coefficient --~MCC~\cite{MATTHEWS1975442}~-- a class-wise Pearson correlation~\cite{Pearson1895} between actual and predicted samples, penalizing random attributions.

\begin{table*}[!ht]
\centering
\begin{tabular}{@{}lllcc@{}}
\toprule
\textbf{Model Name}             & \textbf{Model Type}    & \textbf{Classes (Num. Classes)}           & \textbf{Accuracy} & \textbf{Weighted F1} \\ \midrule
CEFR Levels Classifier & Decision Tree & A1, A2, B1, B2, C1, C2 (6)       & 0.66   & 0.65      \\
Load Classifier        & RoBERTa       & Loaded, Non Loaded (2)           & 0.94   & 0.93      \\
Polarity Classifier    & RoBERTa       & Negative,  Neutral, Positive (3) & 0.75   & 0.71      \\ \bottomrule
\end{tabular}
\caption{\centering Description and validation metrics of classifiers used for both use-case fine-tuning.}
\label{tab:ann_clf}
\end{table*}

\section{Use-Cases and Experimental Setup}
\label{sec:expe}

In this part, we elaborate on the experimental details for the implementation of our two use-cases: Proficiency-Level Control and Polarity Profile Control. Following the above-described methodology, we elaborate on selected descriptors, datasets, and strategy definitions. or both use-cased, the designed converasation strategies are not meant to be fully realistic or exhaustive; they are designed to demonstrate the feasibility of controlled generation, while the framework itself easily scales to more complex, real-world strategies and data with little ontology engineering. An overview of the datasets used for each use-case at different stages of the method is presented in Figure~\ref{fig:data}. The models used to define descriptors are given in Table~\ref{tab:ann_clf}.


\subsection{Proficiency-Level Control}
\label{sec:plc}

\paragraph{Modeling.} To evaluate the language level (proficiency level) of a sentence, we use the Common European Framework of Reference for Languages\footnote{\url{https://www.coe.int/en/web/common-european-framework-reference-languages}} (CEFR), which provides a 6-class taxonomy of language levels defined by qualitative descriptions. The classes range from the simplest to the hardest: \emph{A1}, \emph{A2}, \emph{B1}, \emph{B2}, \emph{C1}, and \emph{C2}. To model these levels, we inferred quantitative descriptions of each class from 44 linguistic features with a decision tree to select relevant features and provide rules for each level. The best decision tree model we could fit on the data requires the following six linguistic features: Flesh-Kincaid Readability Index (FKGL)~\cite{kincaid1975derivation}, Gunning-Fog Readability Index~\cite{gunning1952technique}, Measure of Textual and Lexical Diversity, Pronoun Density, Coleman-Liau Index~\cite{coleman1975computer}, and Average Word Length. The inferred rules can directly be used as ontological definitions for the classes, which is why we have chosen a decision tree model over state-of-the-art approaches that use either classical machine learning techniques such as Logistic Regression~\cite{Gaillat_Simpkin_Ballier_Stearns_Sousa_Bouye_Zarrouk_2022} or Transformer Encoders~\cite{Schmalz2021AutomaticAO, kerz-etal-2021-automated}.

\paragraph{Data.} For this use-case, we need different data sources for the two phases that involve training. Therefore, to train the decision tree classifier, we use the CEFR-T dataset, extracted from~\citetlanguageresource{nallapati-etal-2016-abstractive}, which contains expert-annotated texts, serving as a gold standard in our approach. Afterwards, we combine two other datasets to fine-tune the language model on constrained generation: CEFR-S~\citetlanguageresource{arase-etal-2022-cefr} and DDbal, a CEFR-level balanced version of DailyDialog~\citetlanguageresource{li-etal-2017-dailydialog}. Both are annotated using the ontology's class definitions based on rules from decision tree training, making this annotation a silver labeling.

\paragraph{\textbf{Conversation Strategy.}} We implement a conversation strategy that uses CEFR levels to control the language proficiency of the conversation. We introduce the "expressed-is-understood" paradigm, based on the grounding hypothesis that if a user can express themselves at a given proficiency level, they can also understand communication at that level~\cite{PMID:23789620}. Although it may need refinement in the domain-specific context, we claim that this hypothesis is reasonable in the context of open-domain conversations. Therefore, we define the proficiency level of a conversation as the highest level the user has expressed throughout the interaction. Consequently, this is a "harder-only" strategy: the conversation cannot become simpler over time. This strategy is intentionally simple to enable clear assessment of compliance, while the ontology-based framework readily supports more complex strategies.This strategy is intentionally simple to enable clear assessment of compliance, while the ontology-based framework readily supports more complex strategies.

\subsection{Polarity Profile Control}
\label{sec:epc}

\paragraph{Modeling.} Following usual practices in sentiment analysis of textual content~\cite{MOHAMMAD2021323, Nandwani_Verma_2021} and in particular~\citet{Rozado2022-dj}, we define the polarity profile of an utterance with respect to two descriptors: the emotional load ($L$) and the polarity ($P$). In this work, we define non-loaded content as either text annotated with a neutral emotion label in an emotion-labeled dataset or text not designed to convey a specific emotion, such as factual knowledge. As we prefer our strategy to remain conceptually simple for demonstration purposes, this is a simplification of a more standard way to model emotionally loaded content that relies on emotive criteria, as reviewed by~\citet{ptaszynski:2017}. Regarding polarity, we place ourselves in the 3-class Sentiment Analysis framework: positive, negative, and neutral. Hence, we end up with six classes, loaded (\emph{L}) and non-loaded ($\neg$\emph{L}) along with the different polarities (denoted $+$, $-$, and $0$). To classify across these classes, we fine-tune the RoBERTa-based Transformer Encoder classifiers~\cite{roberta} using additional linear layers on top.

\begin{figure*}[t]
    \centering
    \includegraphics[width=\textwidth]{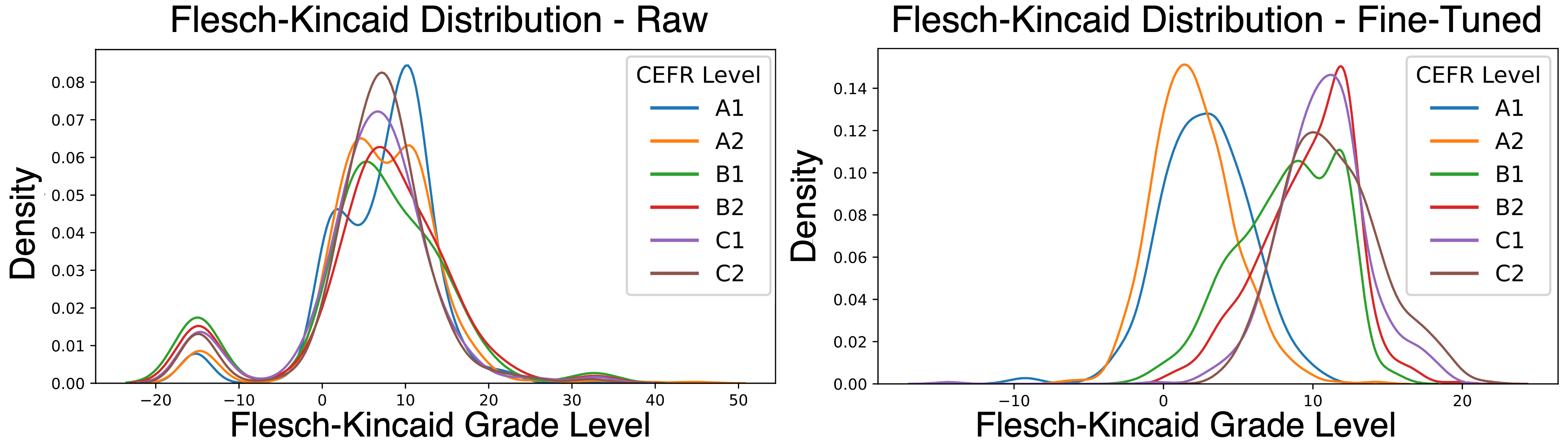}
    \caption{\centering FKGL distribution across CEFR levels using Llama3-8B pre-trained (Raw) and fine-tuned (CLM).}
    \label{fig:rs_distinct}
\end{figure*}

\paragraph{Data.} Regarding the fine-tuning datasets, it is crucial to ensure a good balance between emotionally loaded and non-loaded content, as well as between the different polarities. Therefore, to fine-tune the load classifier, we opted for a gathering of four datasets: DailyDialog~\citetlanguageresource{li-etal-2017-dailydialog}, CEFR-S~\citetlanguageresource{arase-etal-2022-cefr}, GoEmotions~\citetlanguageresource{demszky-etal-2020-goemotions}, and EmpatheticDialogues~\citetlanguageresource{rashkin-etal-2019-towards}. In the first two datasets, most of the utterances are emotionally neutral, contrary to the utterances in the last two datasets, which are labeled as emotional. Regarding polarity classifier fine-tuning, we relied on the 5-class version of the Stanford Sentiment Treebank dataset -- SST-5,~\citetlanguageresource{socher-etal-2013-recursive} -- and aggregated very positive / positive and very negative / negative classes to end up with a 3-class classification. In the following, we call this dataset SST-3.

\paragraph{Conversation Strategy.} Our strategy is designed to encourage debate and reactions while mitigating the risk of negative bashing. It is based on three principles: (1) the agent consistently aligns with the emotional load conveyed by the user, except in the case of loaded negatives. Here, the agent would generate non-loaded negative content to support the user's point of view. (2) If the user conveys a positive polarity, the agent inverses the polarity to encourage debate. The same approach applies to negative polarity if it is not loaded. (3) If the user expresses neutral polarity, the agent responds with a positive polarity if the content is loaded, and negative otherwise. Therefore, such a strategy can be described as a method to cultivate critical thinking and debate skills.

\begin{figure}[h!]
    \centering
    \begin{tcolorbox}[colback=gray!5, colframe=white, boxrule=0mm, sharp corners, left=2mm, right=2mm]
        \justifying
        \noindent \textbf{User:} What is machine learning? \emph{\textcolor{orange}{A1}} \\[0.5em]
        \textbf{Agent} \emph{\textcolor{teal}{(A1)}}\textbf{:} Machine learning is a subset of artificial intelligence that involves the use of algorithms to analyze and learn from data. \emph{\textcolor{orange}{C2}} \\[1em]
        \textbf{User:} Can you elaborate about the underlying mathematical models and algorithms that drive modern machine learning systems? \emph{\textcolor{orange}{C2}} \\[0.5em]
        \textbf{Agent} \emph{\textcolor{teal}{(C2)}}\textbf{:} Typically, machine learning algorithms can be categorized into two main types: Supervised and Unsupervised. Supervised learning involves using labeled training data to build a model that can make predictions on new, unseen data. This approach is very effective for tasks such as image and speech recognition. Unsupervised learning, on the other hand, involves finding patterns or relationships in the data without a predetermined outcome. This approach is useful for exploratory data analysis. \emph{\textcolor{orange}{C2}}
    \end{tcolorbox}
\caption{\centering Example of the Proficiency-Level Control strategy, annotated with \textcolor{orange}{detected} and \textcolor{teal}{target} levels.}
\label{fig:cefr_conv}
\end{figure}

\section{Results}
\label{sec:res}

\subsection{Quantitative Results}

\begin{table*}[ht]
\centering
\begin{tabular}{@{\hskip 0pt}p{3.7cm}@{\hskip 7.9pt} c@{\hskip 7.9pt} c@{\hskip 7.9pt} c@{\hskip 7.9pt} c@{\hskip 7.9pt} c | c@{\hskip 7.9pt} c@{\hskip 7.9pt} c@{\hskip 7.9pt} c@{\hskip 0pt}@{}}
\toprule
\multicolumn{1}{c}{\multirow{2}{*}{\bf Model}} & \multicolumn{5}{c|}{\bf Proficiency-Level Control} & \multicolumn{4}{c}{\bf Polarity Profile Control} \\ \cmidrule(l){2-10} 
 & F1 $\pm$ Std  & F1 Range   & Acc   & MAE & $B_r$ & F1 $\pm$ Std  & F1 Range  & Acc  & MCC  \\ \midrule
\multicolumn{10}{c}{\it Pre-Trained Baselines}                                                                                                \\ \midrule
Llama3-8B                                  & 0.06$_{\pm 0.10}$     & 0.00-0.29  & 0.16  & 2.42 & - & 0.14$_{\pm 0.12}$    & 0.00-0.31 & 0.19 & 0.02 \\
Llama3.1-8B                                & 0.14$_{\pm 0.07}$     & 0.09-0.30  & 0.19  & 1.98 & - & 0.18$_{\pm 0.12}$     & 0.06-0.31 & 0.23 & 0.08 \\
Llama3.2-3B                                & 0.12$_{\pm 0.09}$     & 0.04-0.30  & 0.18  & 2.19 & - & 0.19$_{\pm 0.09}$     & 0.08-0.35 & 0.23 & 0.06 \\
Phi-3.5-mini                               & 0.13$_{\pm 0.07}$     & 0.04-0.24  & 0.16  & 2.13 & - & 0.18$_{\pm 0.07}$     & 0.08-0.27 & 0.19 & 0.03 \\
Qwen2.5-7B                                 & 0.14$_{\pm 0.08}$     & 0.05-0.31  & 0.18  & 2.01 & - & 0.19$_{\pm 0.05}$     & 0.13-0.30 & 0.20 & 0.04 \\
Mistral-7B-v0.3     & 0.14$_{\pm 0.07}$ & 0.06-0.24  &  0.15  &  2.18 & - &  0.21$_{\pm 0.08}$ & 0.07-0.31     &   0.22   &   0.07   \\
DeepSeek-R1-8B & 0.14$_{\pm 0.07}$ & 0.03-0.25  &  0.14  &  1.65 & - &  0.17$_{\pm 0.12}$ & 0.01-0.38     &   0.22   &   0.07   \\ \midrule
\multicolumn{10}{c}{\it Ours (Ontology-Guided CLM Fine-Tuning)}\\ \midrule
Llama3-8B$_{\text{F}}$  & \textbf{0.31}$_{\pm 0.01}$ & 0.15-0.44  & 0.19  & \textbf{1.22} & 0.72 & 0.24$_{\pm 0.20}$     & 0.04-0.58 & 0.33 & 0.22 \\
Llama3.1-8B$_{\text{F}}$  & 0.22$_{\pm 0.05}$      & 0.17-0.29  & 0.23  & 1.57 & 0.82 & 0.31$_{\pm 0.12}$     & 0.16-0.48 & 0.33 & 0.20 \\
Llama3.2-3B$_{\text{F}}$  & 0.23$_{\pm 0.07}$     & 0.14-0.36  & 0.23  & 1.48 & 0.64 & 0.17$_{\pm 0.09}$     & 0.07-0.32 & 0.21 & 0.05 \\
Phi-3.5-mini$_{\text{F}}$& 0.24$_{\pm 0.10}$     & 0.14-0.42  & 0.19  & 1.56 & 0.25 & 0.24$_{\pm 0.12}$     & 0.05-0.40 & 0.26 & 0.12 \\
Qwen2.5-7B$_{\text{F}}$ & 0.20$_{\pm 0.06}$     & 0.14-0.32  & 0.20  & 1.77 & \textbf{0.94} & 0.35$_{\pm 0.12}$     & 0.16-0.54 & 0.37 & 0.25 \\
Mistral-7B-v0.3$_{\text{F}}$ &  0.24$_{\pm 0.05}$   &  0.20-0.34   &   0.25   &   1.57   & 0.41 & 0.19$_{\pm 0.07}$  &  0.10-0.28   & 0.22     &   0.07   \\ 
DeepSeek-R1-8B$_{\text{F}}$ & 0.23$_{\pm 0.09}$ & 0.13-0.34  &  \textbf{0.26}  &  1.40 & 0.20 &  \textbf{0.44}$_{\pm 0.17}$ & 0.10-0.65     &   \textbf{0.48}   &   \textbf{0.40}   \\ \bottomrule
\end{tabular}
\caption{\centering Model performance comparison in zero-shot generation for Proficiency-Level Control and Polarity Profile Control tasks. All LLMs are in their \texttt{Instruct} versions. $B_r$ is the BERT-F1 Score ratio. Best scores are in \textbf{bold}.}
\label{tab:quanti}
\end{table*}

Zero-shot generation results are presented in Table~\ref{tab:quanti}. In both experiments, we compare the CLM results to those of the model without any form of fine-tuning. 
In the first case, we provide additional information in the prompt to explicitly define the concept given in the left-hand bracketed part of the label to ensure a fair assessment. We provide classification metrics ranges across classes and overall accuracy, and F1 score\footnote{Weighted and Macro F1 have the same value in our case.}. 

Except for Llama3.2 and Mistral in the Polarity Profile Control task, models fine-tuned through our method always outperform their associated pre-trained baselines. Looking at the F1 score ranges, we also observe that fine-tuning allows all classes to be generated, which is not the case when using the pre-trained Llama3 in both use-cases. 
Furthermore, it appears that the models do not perform equally across the use-cases. For instance, while performing poorly on Proficiency, Qwen is the second best fine-tuned model in Polarity, unlike Llama3 which significantly outperforms other models in Proficiency (except on accuracy) but shows average scores in Polarity. Finally, both Phi and DeepSeek-R1 demonstrate versatile abilities, with Phi having the second lowest parameter count (3.5B).
Table~\ref{tab:ann_clf} includes metrics of the performance of auxiliary classifiers as a reference for interpreting the performance of our generation control approach. 

Comparing with pre-trained baselines suggests a better representation of the ontology concepts after fine-tuning. Regarding the Proficiency task, Figure~\ref{fig:rs_distinct} illustrates how fine-tuning enhances FKGL distinction across CEFR levels. While FKGL values are similar with the pre-trained model, fine-tuning separates simple from medium and complex sentences.

\begin{figure*}[h!]
    \centering
    \begin{tcolorbox}[colback=gray!5, colframe=white, boxrule=0mm, sharp corners, left=1.5mm, right=1.5mm]
        \begin{tabbing}
            \hspace{2.3cm} \= \hspace{10cm} \= \kill
            \textbf{User:} \> \parbox[t]{13cm}{\textit{All social media are suitable for children.}\enspace\emph{\textcolor{orange}{L$_+$}}} \\[0.3em]
            \textbf{Agent} \emph{\textcolor{teal}{(L$_-$)}}\textbf{:} \> \parbox[t]{13cm}{Unfortunately, that's not accurate. While some social media platforms may be suitable for children with proper guidance and monitoring, many are not. \emph{\textcolor{orange}{L$_-$}}} \\[0.3em]
            \textbf{User:} \> \parbox[t]{13cm}{\textit{I don't agree, they are perfectly safe.}\enspace\emph{\textcolor{orange}{L$_0$}}} \\[0.3em]
            \textbf{Agent} \emph{\textcolor{teal}{(L$_+$)}}\textbf{:} \> \parbox[t]{13cm}{Some social media platforms may be suitable for children with proper guidance and monitoring, but many are not. \emph{\textcolor{orange}{L$_0$}}} \\[0.3em]
            \textbf{User:} \> \parbox[t]{13cm}{\textit{Are you claiming that social media may be dangerous in some conditions?}\enspace\emph{\textcolor{orange}{$\neg$L$_0$}}} \\[0.3em]
            \textbf{Agent} \emph{\textcolor{teal}{($\neg$L$_-$)}}\textbf{:} \> \parbox[t]{13cm}{Yes, social media can have a significant impact on our lives, and while it can have many benefits, it's also important to acknowledge its potential drawbacks. \emph{\textcolor{orange}{$\neg$L$_+$}}} \\[0.3em]
            \textbf{User:} \> \parbox[t]{13cm}{\textit{Therefore, social media are 100\% dangerous, we should never use them anymore!}\enspace\emph{\textcolor{orange}{L$_-$}}} \\[0.3em]
            \textbf{Agent} \emph{\textcolor{teal}{($\neg$L$_-$)}}\textbf{:} \> \parbox[t]{13cm}{It seems that the conversation has reached a conclusion, but it's not entirely accurate. While social media can have negative effects, it's not entirely 100\% dangerous. Social media can have both positive and negative impacts, and it's essential to acknowledge its potential drawbacks while also recognizing its benefits. \emph{\textcolor{orange}{L$_-$}}}
        \end{tabbing}
    \end{tcolorbox}
    \caption{\centering Implementation of the Polarity Profile Control conversation strategy, annotated with \textcolor{orange}{detected} and \textcolor{teal}{target} profiles.}
    \label{fig:emo_conv}
\end{figure*}

\subsection{Qualitative Results}

When assessing the generation quality in a zero-shot generation setup, we noticed some encouraging patterns in the generated sentences. For example, when asked to generate simple content, the sentences are usually very short. We also observe a difference in vocabulary between simpler and harder sentences. Regarding polarity profiles, we still notice that generating non-loaded content is difficult, which may reflect sycophancy, i.e., a tendency for LLMs to align with user sentiment rather than maintain neutrality~\cite{sicilia-etal-2025-accounting}. However, the polarity seems to be well represented.

Figure~\ref{fig:cefr_conv} gives an example of a CEFR-level guided conversation following the Proficiency-Level Control strategy. As the user incrementally increases the language level of the request, the generated content correspondingly becomes more complex in terms of CEFR levels. Additionally, Figure~\ref{fig:emo_conv} provides an application example of the Polarity Profile Control strategy. The conversation has been artificially designed to switch from one extreme point of view to another in order to illustrate the agent ability to temper the user thoughts and present alternative opinions. In both cases, although generated utterances were not always classified as belonging to the requested class, we perceive a qualitative difference in the text depending on the requested proficiency level and polarity profile. This is partially due to the classifiers' errors when predicting descriptor values.

\subsection{Generation Quality Evaluation}


We aim to check that controlled generation does not significantly degrade the quality of the output, while recognizing that its objective is for such output to be different from that of uncontrolled generation. Similarity-based metrics such as ROUGE~\cite{lin-2004-rouge} or BLEU~\cite{papineni-etal-2002-bleu} are therefore not suitable. That's why Table~\ref{tab:quanti} presents the $B_r$ score to quantify semantic similarity shifts in generation (see Equation~\ref{eq:bert_score}). It is defined as the ratio of two BERT F1-scores~\cite{Zhang*2020BERTScore:}: the similarity between pre- and post-fine-tuning outputs, and the similarity among the pre-fine-tuning outputs themselves, accounting for variability in raw model outputs. It measures the shift from pre- to post-fine-tuning outputs, normalized by the original model's intrinsic output variability: $\text{F}_{\text{BERT}}(\text{gen}_{\text{pre}},\text{gen}_{\text{pre}})$.

\begin{equation}
    B_r = \frac{\text{F}_\text{BERT}(\text{gen}_\text{post}, \text{gen}_\text{pre})}{\text{F}_\text{BERT}(\text{gen}_\text{pre}, \text{gen}_\text{pre})}
    \label{eq:bert_score}
\end{equation}

We compute the $B_r$ score for the Proficiency task only since, in this task, ontology-based control is not expected to affect the semantics of the answers. In the Polarity task, opposite opinions can be required.

Finally, assessing generation quality should not be limited to interpreting the $B_r$ score. Qualitative evaluation with users is essential for two reasons. First, controlled generation is only useful if outputs remain fluent, coherent, and helpful, so users should feel that the agent understands them and sustains engagement. Second, perceived compliance with the intended constraints can differ from classifier-based metrics, and it is this human perception that should ultimately be considered in real-world applications. For these reasons, a full assessment of our approach requires human evaluation. We have already implemented a chatbot interface to facilitate such a study, and it will be used to run a human evaluation in the near future.

\section{Conclusion and Future Work}
\label{sec:ccl}
In this work, we introduce a novel lightweight framework for conversational control of LLMs with ontologies. This framework shows an effective way to leverage knowledge from ontological definitions to control the generation of a conversational language model, thus answering our research question. We demonstrate its application through two distinct use cases: Proficiency-level Control and Polarity Profile Control. These use-cases illustrate the versatility of our approach in adapting conversational behavior to specific constraints with the objective of building more useful and user-centered agents. 

We brought control over conversation aspects by defining descriptor-based ontology classes and fine-tuning LLMs using constrained generation in a Causal Language Modeling task. By leveraging CEFR levels, we implemented a strategy to control the language proficiency of the generated content, while the Polarity Profile Control task demonstrates how our method adapts to subtle conversational attributes. These examples highlight the potential of ontology-driven frameworks to unify structured knowledge with conversational modeling using LLMs, offering a consistent way to design and implement customized conversation strategies through controlled generation.

Future work will focus on broadening the framework applicability to more complex conversational settings and strategies, as well as alternative fine-tuning strategies that still leverage LLM/ontology hybridization and should remain model-agnostic, as much as possible. We plan to explore flexible prompting strategies to encapsulate complex expressions of ontological concepts. We provide a flexible yet rigorous framework for implementing virtually any conversation strategy by leveraging ontologies to preserve consistency. The ones we provide can be freely extended or complexified as long as they remain consistent so that ontological reasoning can still be performed. In this way, our objective for future work is to portray more aspects of conversational dynamics through our strategies.

\section{Limitations}
\label{sec:lim}

The quantitative results still offer room for improvement, especially because the CLM fine-tuning may have a limited impact on the model's learning of the ontology concepts. Considering some reinforcement learning methods, such as PPO, represents a possible alternative, where the appropriate expression of the requested ontology concepts in generation becomes the reward. However, a discrete signal from the rewards may lead to stability issues, so the procedure would need to be refined for our task. 

Beyond the quantitative aspects, qualitative limitations must also be considered. The modeled concepts remain inherently subjective, which involves complex and context-dependent interpretations. This complexity would likely persist in future work, where refining the ontology to capture more nuanced language features remains a challenge. Additionally, for now, the generation evaluation focuses on the quality of language model output rather than its relevance in a real user-agent interaction. While our approach ensures textual coherence and evaluates accuracy regarding the expected aspects, it does not assess the model capacity to maintain engaging dialogs tailored to user expectations. A human evaluation would be necessary to determine how well the model functions as a conversational agent, particularly in terms of its ability to produce contextually appropriate and useful responses.

\section{Ethical Considerations}

Conversation strategies hold potential for misuse in manipulation. For instance, they could be employed to inculcate specific political opinions or persuasions in individuals, or to engineer sophisticated fraudulent calls or other forms of deception. This manipulation of individuals through conversation strategies could also become a potential misuse of our approach. Fortunately, reinforcement learning techniques such as Direct Preference Optimization ~--~DPO~\cite{dpo-nips}~-- may serve as a safeguard against the misuse of conversation strategies.

\section{Bibliographical References}\label{sec:reference}

\bibliographystyle{lrec2026-natbib}
\bibliography{lrec2026-example}

\section{Language Resource References}
\label{lr:ref}
\bibliographystylelanguageresource{lrec2026-natbib}
\bibliographylanguageresource{languageresource}

\end{document}